\documentclass[10pt,twocolumn,letterpaper]{article}

\usepackage{times}
\usepackage{epsfig}
\usepackage{graphicx}
\usepackage{amsmath}
\usepackage{amssymb}
\usepackage{booktabs}
\usepackage{multirow}
\usepackage[table, dvipsnames]{xcolor}

\usepackage{iccv}
\usepackage[accsupp]{axessibility}
\definecolor{lightgreen}{rgb}{0.6, 0.8, 0.6}
\definecolor{darkgreen}{rgb}{0.0, 0.5, 0.0}

\newcommand{\increase}[1]{
	{\fontsize{7pt}{0.5em}\selectfont\color{purple}{$\uparrow$~{#1}}}
}
\newcommand{\decrease}[1]{
	{\fontsize{7pt}{0.5em}\selectfont\color{darkgreen}{$\downarrow$~{#1}}}
}

\definecolor{iccvblue}{rgb}{0.21,0.49,0.74}
\usepackage[pagebackref,breaklinks,colorlinks,allcolors=iccvblue]{hyperref}

\def\paperID{3816} 
\def\confName{ICCV}
\def\confYear{2025}

\title{DeRIS: Decoupling Perception and Cognition for Enhanced Enhanced Referring Image Segmentation through Loopback Synergy}

\author{
  Ming Dai$^{1,2\star}$ \ \ Wenxuan Cheng$^{1\star}$, \ \ Jiang-jiang Liu$^2$ \ \ Sen Yang$^2$ \ \ \\ 
  Wenxiao Cai$^3$ \ \ Yanpeng Sun$^2$ \ \ Wankou Yang$^{1\dagger}$ \\
  \\[-6pt]
  $^1$Southeast University \ \ $^2$Baidu VIS \ \ $^3$Stanford University \ \
  \\[-3pt]
  \small
  $\star$ Equal contribution. \quad $\dagger$ Corresponding author.
}

\begin{document}
\maketitle

\begin{abstract}
  Referring Image Segmentation (RIS) is a challenging task that aims to segment objects in an image based on natural language expressions. While prior studies have predominantly concentrated on improving vision-language interactions and achieving fine-grained localization, a systematic analysis of the fundamental bottlenecks in existing RIS frameworks remains underexplored. To bridge this gap, we propose \textbf{DeRIS}, a novel framework that decomposes RIS into two key components: \textit{perception} and \textit{cognition}. This modular decomposition facilitates a systematic analysis of the primary bottlenecks impeding RIS performance. Our findings reveal that the predominant limitation lies not in perceptual deficiencies, but in the insufficient multi-modal cognitive capacity of current models. To mitigate this, we propose a Loopback Synergy mechanism, which enhances the synergy between the perception and cognition modules, thereby enabling precise segmentation while simultaneously improving robust image-text comprehension. Additionally, we analyze and introduce a simple non-referent sample conversion data augmentation to address the long-tail distribution issue related to target existence judgement in general scenarios. Notably, \textbf{DeRIS} demonstrates inherent adaptability to both non- and multi-referents scenarios without requiring specialized architectural modifications, enhancing its general applicability. The codes and models are available at \href{https://github.com/Dmmm1997/DeRIS}{link}

\end{abstract}

\section{Introduction}

\begin{figure}
  \centering
  \includegraphics[width=1.0\linewidth]{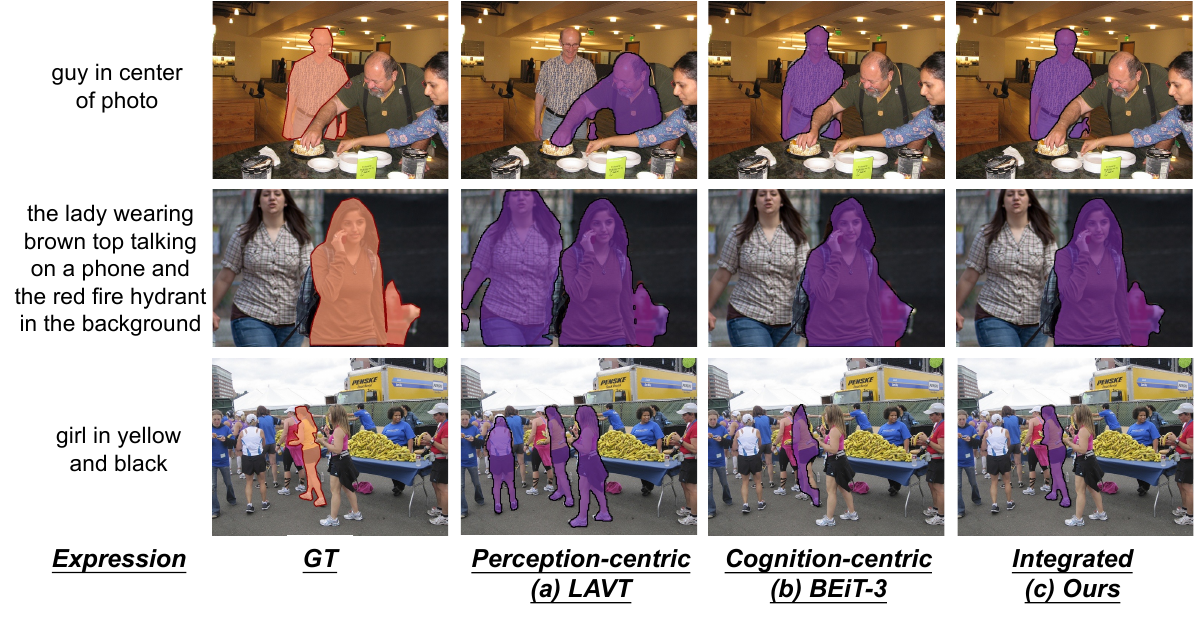}
  \vspace{-20pt}
  \caption{The visualizations of architectures with different emphases are shown in Fig.~\ref{fig:motivation}. (a) The \textit{perception-centric} structure emphasizes fine-grained hierarchical features. (b) The \textit{cognition-centric} structure comprehensively understands the image-text context. (c) Our DeRIS integrates the advantages of both.}
  \vspace{-10pt}
  \label{fig:motivation_visual_improve}
\end{figure}

\begin{figure*}
  \centering
  \includegraphics[width=0.95\linewidth]{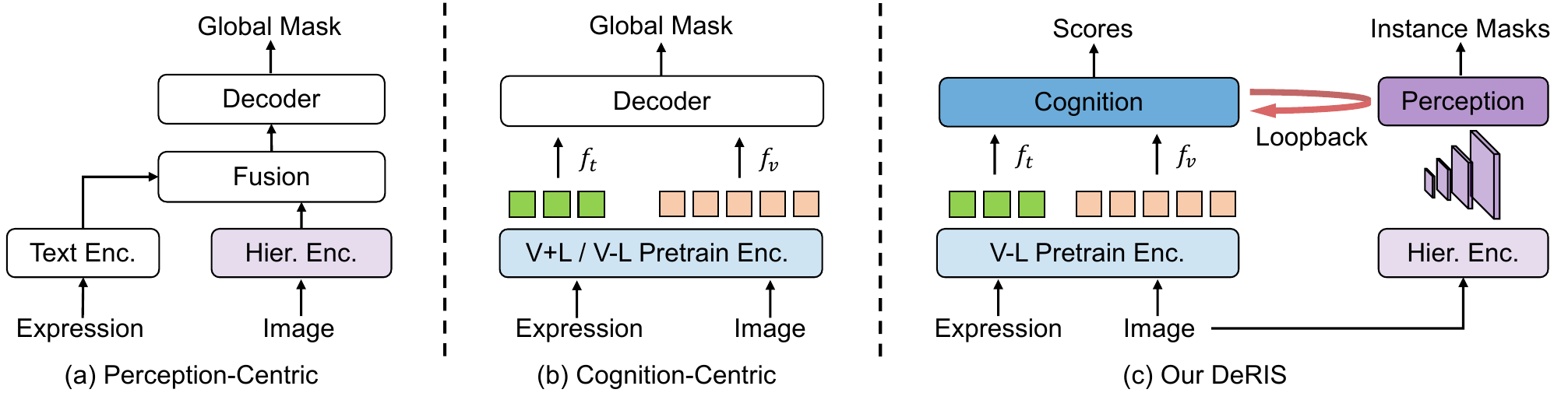}
  \vspace{-10pt}
  \caption{Comparison of architectural paradigms:  
  (a) \textit{Perception-centric} models rely on hierarchical encoders~\cite{resnet,yolov3,swin} to preserve fine-grained spatial features.  
  (b) \textit{Cognition-centric} models leverage vision-language pre-trained model to enhance multi-modal representation and alignment, where V+L refers to two-stream models~\cite{clip,regionclip, declip}, and V-L denotes one-stream models~\cite{beit3,vilt}.  
  (c) The proposed \textit{DeRIS} framework, which integrates robust cognition and fine-grained perception capabilities.}
  \vspace{-10pt}
  \label{fig:motivation}
\end{figure*}

Referring Image Segmentation (RIS)~\cite{hu2016segmentation, li2018referring, liu2017recurrent} is a fundamental yet challenging task in multi-modal understanding. Unlike traditional segmentation tasks~\cite{maskrcnn, mask2former, maskdino}, which are restricted to predefined and fixed object categories, RIS offers greater flexibility by enabling object segmentation based on free-form textual descriptions. This requires a deeper alignment between diverse linguistic expressions and the visual context.  
Early RIS tasks~\cite{refcoco, refcocog, refcocog-umd} typically assumed a one-to-one correspondence between language expressions and target objects. To improve generalization, recent studies~\cite{rela, rris, refzom} have introduced more generalized formulations, enabling segmentation of multiple targets or handling non-referent cases.

In this paper, we focus on the core capabilities reflected by model architectures and categorize mainstream RIS methods into two types: \textit{perception-centric} and \textit{cognition-centric}. As illustrated in Fig.~\ref{fig:motivation}(a), perception-centric methods~\cite{lavt,vlt,cris,seqtr,polyformer,pvd} rely on hierarchical backbones~\cite{resnet,swin,yolov3} designed for perceptual tasks, which effectively preserve fine-grained spatial information. However, due to the limited diversity in downstream datasets, multi-modal fusion modules, when trained from scratch, often exhibit weaker content cognition capabilities. As shown in Fig.~\ref{fig:motivation_visual_improve}(a), while perception-centric methods excel at retaining fine-grained contour information, they struggle with the contextual understanding of image-text relationships.
In contrast, cognition-centric methods~\cite{shareRIS,simvg,oneref,c3vg} (structure is illustrated in Fig.~\ref{fig:motivation}(b)) leverage large-scale vision-language pre-trained models~\cite{clip,vilt,vlmo,beit3} to significantly enhance multimodal understanding and alignment capabilities. However, due to the transformer architecture employed, which incurs quadratic computational complexity with respect to input resolution, these methods suffer from a notable limitation: a substantial loss of fine-grained spatial information. This deficiency undermines their performance in high-precision perception tasks. As shown in Fig.~\ref{fig:motivation_visual_improve}(b), while cognition-centric approaches can effectively identify corresponding regions, they struggle to delineate precise contour boundaries, highlighting a critical trade-off between comprehension and localization accuracy.

Fundamentally, we argue that RIS inherently involves two critical dimensions: \textit{perception} and \textit{cognition}. Perception focuses on accurately localizing foreground objects, whereas cognition entails a comprehensive understanding of both textual and visual content. Given the significance of both aspects, we propose a novel framework, \textbf{DeRIS}, which explicitly decouples perception and cognition, as illustrated in Fig.~\ref{fig:motivation}(c). This design effectively integrates the strengths of both perception- and cognition-centric paradigms, enabling robust multi-modal understanding while preserving fine-grained perceptual performance. As shown in Fig.~\ref{fig:motivation_visual_improve}(c), DeRIS not only maintains high-precision perception but also enhances contextual comprehension. 
The remainder of this paper primarily discusses the following three issues.

\begin{figure}
  \centering
  \includegraphics[width=1.0\linewidth]{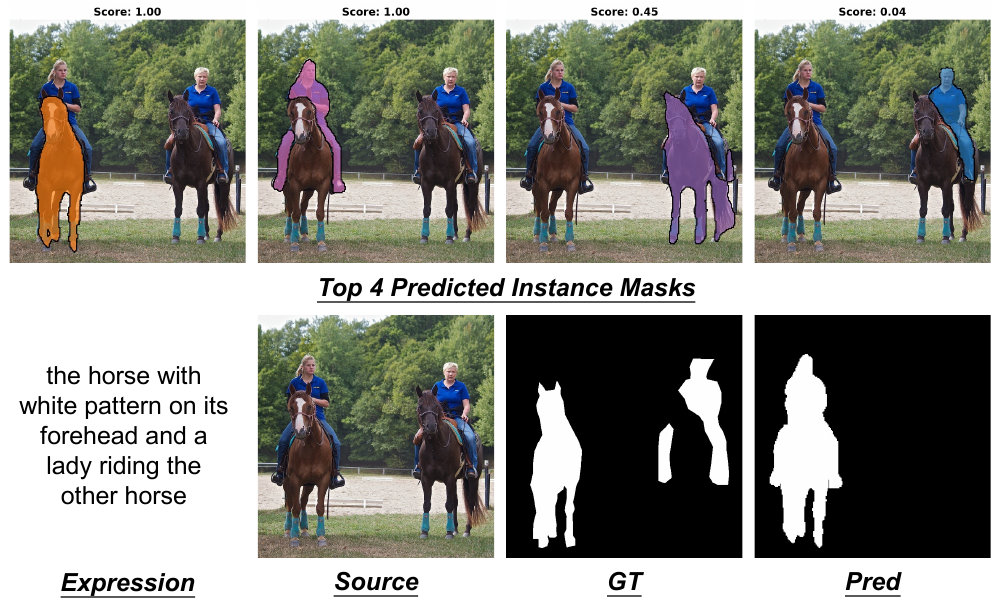}
  \vspace{-20pt}
  \caption{Visualization of strong perception with weak cognition.}
  \vspace{-15pt}
  \label{fig:motivation_weakcognition}
\end{figure}

\textbf{\textit{(1) What constitutes the primary bottleneck in RIS: perception or cognition?}}  
DeRIS decouples perception and cognition to independently analyze their contributions to RIS performance.  
From a quantitative perspective, Table~\ref{tab:ab_recognition_perception} presents the impact of progressively enhancing perception and cognition. Our results show that while improving perception yields only marginal gains (+1.2\% cIoU), enhancing cognition leads to a substantial performance boost (+12.9\% cIoU).
From a qualitative perspective, Fig.~\ref{fig:motivation_weakcognition} visualizes the top 4 instance masks. Although object queries produce accurate masks, referent classification frequently fails due to inadequate contextual and semantic understanding. These findings conclusively demonstrate that \textit{cognition, rather than perception, constitutes the primary bottleneck}. More details are analyzed in Sec.~\ref{sec:ab_cognition_perception}.

\textbf{\textit{(2) How to Effectively Connect Perception and Cognition Components?}}
We propose a loopback architecture to establish strong interactions between the cognition and perception. In this architecture, object queries serve as a medium to generate fine-grained instance-level masks within the perception branch, which are subsequently transferred to the cognition branch to compute referring scores with robust contextual understanding. The process is iteratively reinforced, with the predictions from each iteration being supervised. This progressively information flow forms a powerful loopback synergy, fostering strong connections between the perception and cognition branches. Some synergy forms are compared in Sec.~\ref{sec:ab_loopback}.

\textbf{\textit{(3) How to Improve Non-Referent Judgment Ability?}}  
We analyze the gRefCOCO~\cite{rela} dataset and observe that approximately 9\% of the samples are non-referents, leading to a long-tail distribution problem in non-referent classification. To mitigate this issue, we propose a non-referent sample conversion augmentation strategy that dynamically transforms referent-present expressions into non-referent ones, thereby increasing the diversity of non-referent sample pairs.  
Furthermore, we find that auxiliary supervision for non-referent judgment improves the model's capability to handle non-referent cases, even though it is not explicitly leveraged during testing. We provide quantitative analyses in Sec.~\ref{sec:ab_non_referent} and qualitative analyses in Appendix F.3.

The main contributions of this paper are as follows:
\begin{itemize}
\item We introduce \textbf{DeRIS}, a pioneering framework that decouples the perception and cognition components of RIS, seamlessly integrating their respective strengths to achieve high-precision perceptual localization and robust multi-modal contextual understanding.
\item We provide in-depth analyses of the inherent limitations in existing RIS frameworks, identifying critical structural bottlenecks and the long-tail distribution challenges. To address these issues, we propose a novel \textit{Loopback Synergy} mechanism that fosters progressively interaction between perception and cognition modules. Furthermore, we develop a simple yet effective non-referent sample conversion strategy to alleviate training instability and enhance model generalization.
\item Our proposed approach sets a new state-of-the-art performance on the RefCOCO, RefCOCO+, and RefCOCOg (RIS), as well as the gRefCOCO (GRES) dataset.
\end{itemize}

\section{Related Work}
\label{sec:related_work}

\subsection{Referring Image Segmentation (RIS)}
Perception-centric RIS methods are typically categorized into two structural paradigms: post-fusion and early-fusion. Post-fusion methods~\cite{mcn,reftr,restr,cris,ISF, seqtr,polyformer,huang2023referring} primarily focus on the fusion of visual and linguistic features through specially designed interaction modules that are applied after the visual backbone. In contrast, early-fusion methods~\cite{vlt,efn,lavt,M3Att,qrnet,coupalign,pvd,vg-law,eevg} emphasize integrating visual and linguistic features during the feature extraction phase via modularized architectural designs, facilitating deeper cross-modal understanding.
Cognition-centric RIS methods~\cite{simvg,shareRIS,oneref,c3vg,instancevg, propvg} leverage powerful pre-trained multimodal models~\cite{clip,vlmo,vilt,beit3} to generate aligned or fused multimodal representations, thereby enhancing the model’s cognitive capacity for interpreting complex image-text relationships. In this work, we introduce a novel architectural framework that decouples the perception and cognition components, effectively harnessing the strengths of both paradigms to advance RIS performance.

\subsection{Vision-Language Understanding}
Vision-language pre-trained models can be broadly classified into four primary categories: one-stream, dual-stream, fusion encoder, and multimodal large language models (MLLMs).
One-stream models~\cite{uniter,vilt,alberf,soho} process image and text inputs within a unified stream, concatenating their embeddings to enable cross-modal interactions throughout the feature extraction process.
Dual-stream models~\cite{clip,align,declip,regionclip}, in contrast, employ separate encoders for each modality, with interactions occurring at a shallow layer between pooled image and text features.
Fusion encoder models~\cite{blip,vlmo,flava,glipv2,beit3} integrate elements of both one- and dual-stream approaches, facilitating intermediate cross-modal interactions to balance model complexity and performance.
Recently, MLLM-based methods~\cite{llava,mipha,Qwen2VL,deepseekvl2} harness the advanced reasoning and understanding capabilities of foundational large language models, adapting them for multimodal inputs through tailored training strategies.
In this paper, we identify a key deficiency in current models: \textit{inadequate cognitive capability}. To address this, we introduce a loopback synergy strategy that employs object queries as an intermediary to facilitate cascading and progressive transfers of semantic understanding between the target perception and text-image modalities.

\subsection{Instance Segmentation}
CNN-based instance segmentation methods can be classified into two major paradigms: top-down and bottom-up. The top-down paradigm~\cite{maskrcnn,cascadercnn,instanceasquery,tensormask,blendmask,yolact} first generates bounding boxes using object detectors, followed by mask segmentation. In contrast, the bottom-up paradigm~\cite{newell2017associative,de2017semantic,sgn,ssap} approaches instance segmentation as a labeling and clustering problem, where pixels are first assigned class labels or embedded into a feature space, then clustered into individual instances.
Recently, several transformer-based instance segmentation methods~\cite{oneformer,soit,mask2former,maskdino} have been proposed. These methods treat instance segmentation as a mask classification task, associating instance categories with a set of predicted binary masks.  
In this work, we adopt transformer-based architectures and leverage pre-trained weights from Mask2Former~\cite{mask2former} to provide the model with prior perception capabilities. This design introduces instance-aware capabilities into the RIS task, enabling more refined spatial discrimination of instances.

\begin{figure*}
  \centering
  \includegraphics[width=0.7\linewidth]{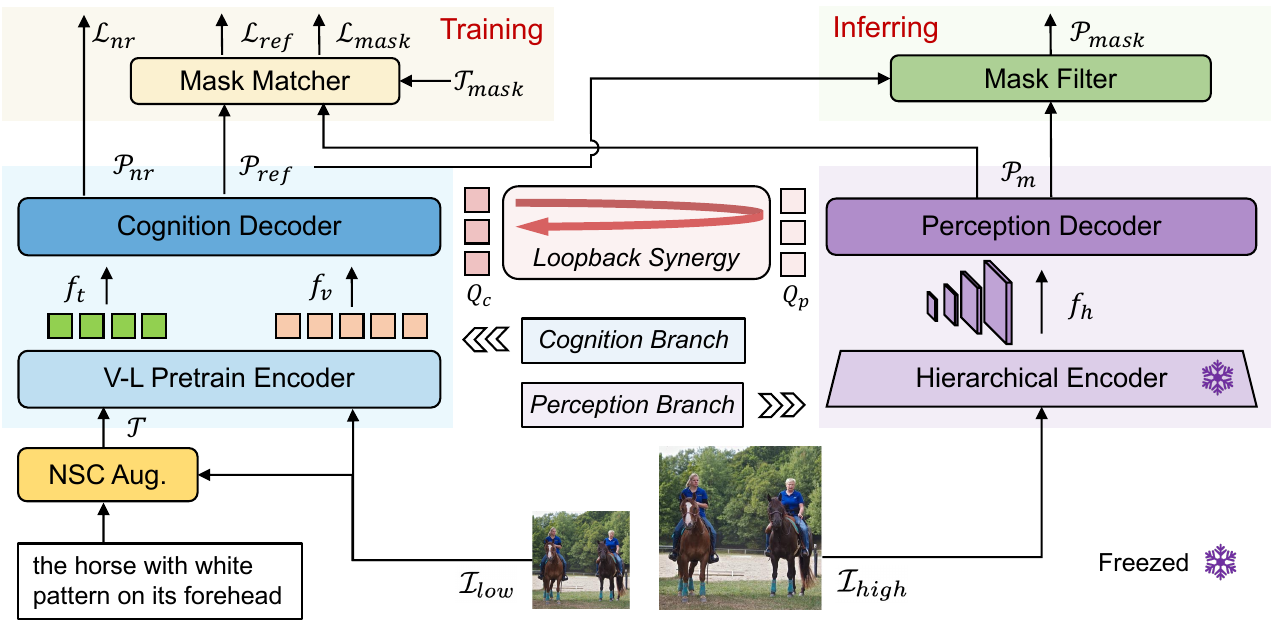}
  \vspace{-5pt}
  \caption{Overview of the proposed DeRIS framework. The RIS task is decoupled into \textit{perception} and \textit{cognition} branches, with a loopback synergy mechanism facilitating iterative information exchange. This design enhances synergy between the two branches, enabling a dynamic and progressive understanding of both perceptual targets and multi-modal semantics.}
  \vspace{-5pt}
  \label{fig:framework}
\end{figure*}

\section{Methodology}

\subsection{Overview}

Fig.~\ref{fig:framework} illustrates an overview of the DeRIS architecture, comprising three primary components: \textit{perception branch}, \textit{cognition branch}, and \textit{loopback synergy}.  
In the cognition branch, we introduce a non-referent sample conversion (NSC) strategy, which dynamically transforms referent image-text pairs into non-referent samples. The low-resolution image $\mathcal{I}_{low}$ and augmented text $\mathcal{T}'$ are then jointly embedded using an image-text encoder, leveraging the  vision-language pre-trained (BEiT3~\cite{beit3}) model. The choice of low resolution is motivated by two factors: (i) mitigating the quadratic computational complexity associated with input resolution, and (ii) the cognition branch primarily focuses on referent classification based on semantic understanding rather than fine-grained perception. 
The perception branch follows a structure similar to Mask2Former~\cite{mask2former}, employing a hierarchical encoder~\cite{swin} to extract multi-scale features from high-resolution images $\mathcal{I}_{high}$. These features are then refined using an FPN-like~\cite{fasterrcnn} structure to enhance fine-grained visual representations.  
The loopback synergy adopts a cascade structure, enabling multiple rounds of object query transfer between the perception and cognition branches. This iterative process fosters a dynamic and progressive understanding of both perceptual targets and multi-modal semantics. By refining multi-level features, the model generates high-quality masks while concurrently improving cognitive capabilities.  
Finally, non-referent samples are identified through query pooling and a binary classifier, formulated as:
\begin{equation}
  \setlength{\abovedisplayskip}{5pt}
  \setlength{\belowdisplayskip}{5pt}
  \mathcal{P}_{nr} = \text{AvgPool}(\text{Linear}(Q_{ck}))
\end{equation}
where $Q_{ck}$ denotes the cognition queries from the final round of loopback synergy.

\subsection{Loopback Synergy}
\subsubsection{Single-round of Loopback Synergy}
\begin{figure*}
  \centering
  \includegraphics[width=1.0\linewidth]{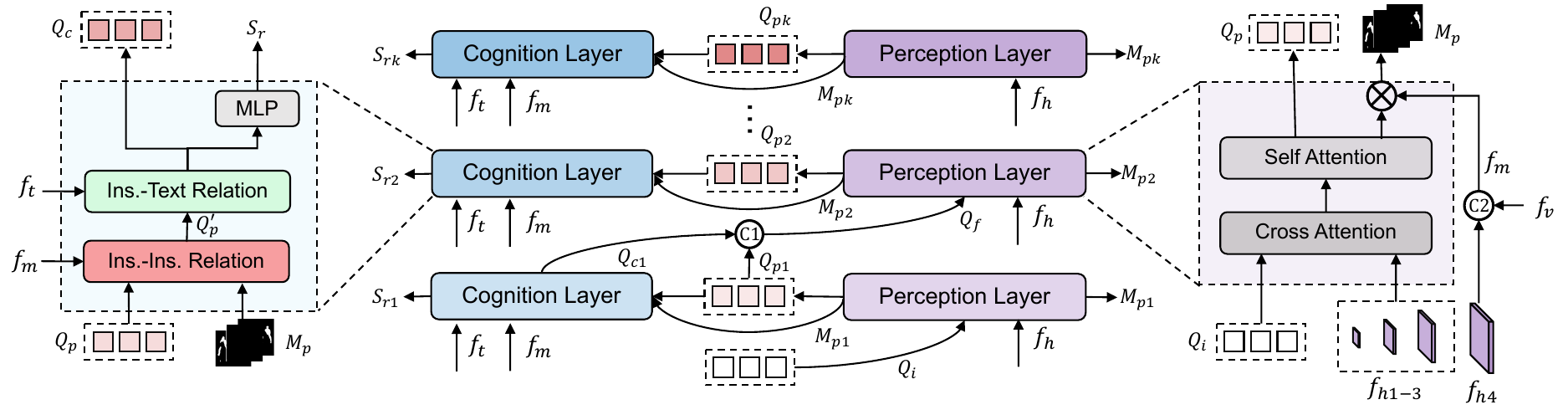}
  \vspace{-20pt}
  \caption{The architecture of the loopback synergy. Each round of interaction consists of a cognition layer and a perception layer. The perception layer provides object queries $Q_p$ and the decoded mask $M_p$ to the cognition layer. The cognition layer interacts with these, and produces cognition queries $Q_c$ and referent scores $S_r$.}
  \vspace{-5pt}
  \label{fig:loopback_arch}
\end{figure*}

The loopback synergy module is designed to establish strong interactions between the perception and cognition branches. As illustrated in Fig.~\ref{fig:loopback_arch}, each round of interaction consists of a \textit{cognition layer} and a \textit{perception layer}. Initially, the perception layer provides perceptual queries $Q_p$ and the decoded mask $M_p$ to the cognition layer. Subsequently, the cognition layer interacts with $Q_p$ and the image-text semantic information, producing cognitive query $Q_c$ and a confidence score $S_r$ for each query, which indicates the relevance to the target object. Finally, the fused query $Q_f$ for the next round is obtained by applying the $\text{C1}$ operation between $Q_p$ and $Q_r$.
The C1 operation is defined as:
\begin{equation}
  \setlength{\abovedisplayskip}{5pt}
  \setlength{\belowdisplayskip}{5pt}
  Q_f = \text{C1}(Q_p, Q_r) = \text{MLP}(\text{Concat}(Q_p, Q_r)).
\end{equation}
This bidirectional information flow forms an effective loopback mechanism, creating robust connections between the perception and cognition branches.

\subsubsection{Perception Layer}
The operations in the perception layer closely resemble those in Mask2Former~\cite{mask2former}, as depicted in the right portion of Fig.~\ref{fig:loopback_arch}. Initially, initial queries $Q_i$ interact with multi-scale features via deformable cross-attention to capture spatial contextual information. This is followed by a self-attention operation to establish inter-instance relationships, yielding the output $Q_p$. Subsequently, the predicted mask $M_p$ is derived by performing a dot product between $Q_p$ and the feature map $f_m$. The feature map $f_m$ integrates both perception-related features $f_{h4}$ and cognition-related features $f_v$, defined as:
\begin{equation}
  \setlength{\abovedisplayskip}{5pt}
  \setlength{\belowdisplayskip}{5pt}
  f_m = \text{C2}(f_{h4}, f_v) = \text{Conv}(\text{Concat}(f_{h4}, f_v)),
\end{equation}
where $\text{Conv}$ denotes a 1$\times$1 convolution operation. This design primarily aims to bolster the model’s capability to generate text-informed proposals by incorporating contextual semantic information from the cognition branch, while simultaneously enhancing its perceptual accuracy.

\subsubsection{Cognition Layer}
The operations in the cognition layer are depicted in the left portion of Fig.~\ref{fig:loopback_arch}. The Instance-Instance Relation aims to establish mutual relationships among $Q_p$, with the goal of enabling each object to perceive its spatial position. We leverage the mask prediction $M_p$ from the perception layer as a prior. The computation proceeds as follows:
\begin{equation}
  \setlength{\abovedisplayskip}{5pt}
  \setlength{\belowdisplayskip}{2pt}
  f_s = \text{AvgPool}(f_m \times \sigma(M_p)),
\end{equation}
\begin{equation}
  \setlength{\abovedisplayskip}{2pt}
  \setlength{\belowdisplayskip}{5pt}
  Q_p' = \text{Attn}(Q_i, Q_i+f_s, Q_i),
\end{equation}
where $\sigma$ denotes the sigmoid operation, and $\text{Attn}$ refers to the self-attention operation. The Instance-Text Relation employs $Q_p'$ as the query and the textual features $f_t$ as both the key and value in a cross-attention operation, yielding $Q_c$, which is closely aligned with image-text semantics. Finally, an MLP is applied to derive the confidence score $S_r$, representing the probability of each query’s relevance to the input expression.

\subsection{Non-referent Sample Conversion}
\label{sec:nsc}
We observe that existing models frequently demonstrate suboptimal performance in non-referent judgment, as evidenced by the N-acc metric in general scenarios. Our analysis reveals that the proportion of non-referent samples in the gRefCOCO~\cite{rela} training set is approximately 9\%, highlighting a long-tail distribution challenge. Due to the limited representation of non-referent samples during training, models tend to over-predict the presence of target objects, leading to reduced N-acc scores. To mitigate this issue, we propose a simple non-referent sample conversion augmentation strategy, which dynamically transforms image-text pairs containing targets into non-referent samples by replacing the textual description, thereby increasing the diversity of non-referent sample pairs. However, this approach requires careful design to prevent scenarios where the described target in the modified text might still exist in the corresponding image. We address this challenge through a three-level filtering process:
\begin{enumerate}
  \item The image corresponding to the selected sentence must be inconsistent with the current image.
  \item The length of the selected sentence should be greater than a threshold $N_w$.
  \item The similarity between the selected sentence and the original sentence should be below a threshold $T_s$.
\end{enumerate}
Here, $N_w$ and $T_s$ are set to 2 and 0.6 by default. The similarity is computed using a fast approximation method:
\begin{equation}
  \setlength{\abovedisplayskip}{5pt}
  \setlength{\belowdisplayskip}{5pt}
  \text{Sim}(s_1, s_2) = \frac{\text{Jac}(s_1, s_2) + \text{Cos}(s_1, s_2)}{2}
\end{equation}
where Jac and Cos denote the Jaccard and Cosine similarity (See Appendix D.1). Additionally, we introduce an extra hyperparameter $R_c$, which controls the probability of conversion (See  Sec.~\ref{sec:ab_R_c}).

\subsection{Training Objectives}
\label{sec:training_objectives}
The output of DeRIS consists of three components: the segmentation mask $\mathcal{P}_m$, referring classification $\mathcal{P}_{ref}$, and non-referent judgment $\mathcal{P}_{nr}$. Initially, a mask matcher~\cite{mask2former} is used to perform Hungarian matching between $\mathcal{P}_m$ and the ground truth mask $\mathcal{T}_{mask}$.
The training objective consists of three components: 
\begin{enumerate}
  \item The segmentation loss $\mathcal{L}_{mask}$, associated with each matched $\mathcal{P}_m$, is computed using a combination of BCE and Dice loss to quantify the discrepancy between the predicted mask and the ground truth mask.
  \item The referring classification loss $\mathcal{L}_{r}$, corresponding to each object query, also employs BCE loss to assess whether the query correctly refers to the target object. 
  \item The non-referent loss $\mathcal{L}_{nr}$ utilizes BCE loss to determine whether the described object exists in the image.
\end{enumerate}
The loss for the $i$-th loopback round can be expressed as in Eq.~\ref{eq:eachround_loss}, and the total loss is given by Eq.~\ref{eq:total_loss}.
\begin{equation}
\label{eq:eachround_loss}
  \setlength{\abovedisplayskip}{5pt}
  \setlength{\belowdisplayskip}{5pt}
  \mathcal{L}^i = \lambda_{m}\mathcal{L}_{mask}^i +  \lambda_{r}\mathcal{L}_{r}^i +  \lambda_{nt}\mathcal{L}_{nt}^i,
  \vspace{-5pt}
\end{equation}
\begin{equation}
  \label{eq:total_loss}
  \setlength{\abovedisplayskip}{5pt}
  \setlength{\belowdisplayskip}{5pt}
  \mathcal{L} = \lambda_{aux}\sum_{i=1}^{N_r-1}{\mathcal{L}^i} + \mathcal{L}^{N_r},
  \vspace{-5pt}
\end{equation}

where $\lambda_{m}$, $\lambda_{r}$, and $\lambda_{nt}$ are set to 1.0, 1.0, and 1.0 by default. $\lambda_{aux}$ controls the weight of the auxiliary losses, which is set to 0.2. $N_r$ denotes the number of interaction rounds, which is set to 3.
During inference, the referring classification $\mathcal{P}_{ref}$ is filtered using a threshold $\mathcal{T}_{ref}$. In this paper, we set $\mathcal{T}_{ref}=0.7$ by default.

\begin{table*}
	\renewcommand{\arraystretch}{1.2}
	\setlength{\tabcolsep}{11pt}
  \centering
  \caption{Comparison with state-of-the-art methods on the RefCOCO/+/g~\cite{refcoco,refcocog-umd} datasets for the RIS task. \textsuperscript{†} indicates the use of the cIoU evaluation metric, \textsuperscript{‡} indicates the use of oIoU, while all other methods are evaluated using mIoU.}
  \vspace{-5pt}
  \resizebox{1.0\linewidth}{!}{
	\begin{tabular}{l|c|c|ccc|ccc|cc}
		\specialrule{.1em}{.05em}{.05em}
		\multicolumn{1}{c|}{\multirow{2}{*}{Method}} & Visual & Textual &
		\multicolumn{3}{c|}{RefCOCO} & \multicolumn{3}{c|}{RefCOCO+} & \multicolumn{2}{c}{RefCOCOg} \\
		\cline{4-11}
		& Encoder & Encoder & val & test A & test B & val & test A & test B & val(U) & test(U) \\
        \hline
        \multicolumn{11}{l}{\textit{\textbf{MLLM-based Methods:}}} \\
        LISA-7B\textsuperscript{†}~[CVPR'24]~\cite{lisa} & SAM-H + CLIP-L & Vicuna-7B & 74.90 & 79.10 & 72.30 & 65.10 & 70.80 & 58.10 & 67.90 & 70.60 \\
		    GSVA-13B\textsuperscript{†}~[CVPR'24]~\cite{gsva} & SAM-H + CLIP-L & Vicuna-13B &78.20 & 80.40 & 74.20 &67.40 &71.50& 60.90& 74.20& 75.60 \\
        GLaMM-7B\textsuperscript{†}~[CVPR'24]~\cite{glamm} & SAM-H + CLIP-H & Vicuna-7B & 79.50 & \underline{83.20} & \underline{76.90} & 72.60 & 78.70 & 64.60 & 74.20 & 74.90 \\
        SAM4MLLM-8B\textsuperscript{†}~[ECCV'24]~\cite{sam4mllm} & SAM-EfficientViT-XL1 & Qwen-VL-7B & \underline{79.80} & 82.70 & 74.70 & \underline{74.60} & \underline{80.00} & \underline{67.20} & \underline{75.50} & \underline{76.40} \\
        \hline
        \rowcolor{gray!10} \textbf{DeRIS-7B\textsuperscript{†} (Ours)} & Swin-B + SigLIP & Qwen2-OV-7B & \textbf{84.05} & \textbf{85.79} & \textbf{83.32} & \textbf{80.30} & \textbf{83.92} & \textbf{76.16} & \textbf{80.62} & \textbf{80.59} \\
        \hline
        \hline
        \multicolumn{11}{l}{\textit{\textbf{Specialist Methods:}}} \\
        PVD~[AAAI'24]~\cite{pvd} & Swin-B & BERT-B  & 74.82 & 77.11 &  69.52 & 63.38 & 68.60 & 56.92 & 63.13 & 63.62 \\
        ReMamber\textsuperscript{‡}~[ECCV'24]~\cite{remamber} & Mamba-B & CLIP-B & 74.54 & 76.74 & 70.89 & 65.00 &70.78 &57.53 &63.90 &64.00\\
        VG-LAW~[CVPR'23]~\cite{vg-law} & ViT-B & BERT-B &75.62 & 77.51 &72.89 &66.63 &70.38 &59.89 &65.53 &66.08 \\
        CGFormer~[CVPR'23]~\cite{cgformer} & Swin-B & BERT-B & 76.93 & 78.70 & 73.32 & 68.56 &73.76 &61.72 &67.57 &67.83 \\
        PolyFormer-L~[CVPR'23]~\cite{polyformer} & Swin-L & BERT-B  & 76.94 & 78.49 & 74.83 & 72.15 & 75.71 & 66.73 & 71.15 &71.17 \\
        Prompt-RIS~[CVPR'24]~\cite{prompt-ris} & SAM-B + CLIP-B & CLIP-B  & 78.10 & {81.21} & 74.64 & 71.13& 76.60 & 64.25 & 70.47 & 71.29 \\
        EEVG~[ECCV'24]~\cite{eevg} & ViT-B & BERT-B  & {79.49} & 80.87 & {77.39} & {71.86} & {76.67} & {66.31} &{73.56} & {73.47}  \\
        SimVG-Seg~[NeurIPS'24]~\cite{simvg} & BEiT3-B & BEiT3-B & 77.78 & 79.14 & 76.02 & 72.21 & 75.37 & 67.85 & 72.19 & 73.02 \\
        C3VG~[AAAI'25]~\cite{c3vg} & BEiT3-B & BEiT3-B & 81.37 & 82.93 & 79.12 & \underline{77.05} & 79.61 & 72.40 & \underline{76.34} & 77.10 \\
        OneRef-L~[NeurIPS'24]~\cite{oneref} & BEiT3-L & BEiT3-L & {81.26} & \underline{83.06}  & {79.45}& 76.60 & \underline{80.16} & \underline{72.95} & {75.68}& {76.82} \\
		\hline
      \rowcolor{gray!10}   \textbf{DeRIS-B (Ours)} & Swin-S & BEiT3-B &\underline{81.99} & {82.97} & \underline{80.14} & {75.62} & {79.16} & {71.63} & 76.30 & \underline{77.15} \\
      \rowcolor{gray!10}   \textbf{DeRIS-L (Ours)} & Swin-B & BEiT3-L & \textbf{85.72} & \textbf{86.64} & \textbf{84.52} & \textbf{81.28} & \textbf{83.74} & \textbf{78.59} & \textbf{80.01} & \textbf{81.32} \\
      \specialrule{.1em}{.05em}{.05em}
	\end{tabular}}
	\label{tab:sota_ris}
  \vspace{-5pt}
\end{table*}

\begin{table*}
  \setlength{\tabcolsep}{9pt}
  \centering
  \renewcommand\arraystretch{1.2}
  \caption{Comparison with the state-of-the-art methods on gRefCOCO~\cite{rela} dataset for GRES task.}
  \vspace{-5pt}
  \resizebox{1.0\linewidth}{!}{
      \begin{tabular}{l|c|c|ccc|ccc|ccc}
      \specialrule{.1em}{.05em}{.05em}
      \multicolumn{1}{c|}{\multirow{2}{*}{Method}} & Visual & Textual& \multicolumn{3}{c|}{Val} & \multicolumn{3}{c|}{TestA}  & \multicolumn{3}{c}{TestB} \\
      \cline{4-12}
      & Encoder& Encoder& gIoU & cIoU & N-acc. & gIoU & cIoU & N-acc. & gIoU & cIoU & N-acc.\\
      \hline
      \multicolumn{12}{l}{\textit{\textbf{MLLM-based Methods:}}} \\
      LISA-7B~[CVPR'24]~\cite{lisa}& SAM-ViT-H & Vicuna-7B & 61.63 & 61.76 & 54.67 & 66.27 & 68.50 & 50.01 & 58.84 & 60.63 & 51.91 \\
      GSVA-7B~[CVPR'24]\cite{gsva}& SAM-ViT-H & Vicuna-7B & 66.47 & 63.29 & 62.43 & 71.08 & 69.93 & 65.31 & 62.23 & 60.47 & 60.56 \\
      SAM4MLLM-8B~[ECCV'24]~\cite{sam4mllm}& SAM-EfficientViT-XL1  & Qwen-VL-7B & 71.86 & 67.83 & 66.08 & \underline{74.15} & \underline{72.22} & 63.92 & 65.29 & 63.42 & 59.99 \\
      \hline
      \hline
      \multicolumn{12}{l}{\textit{\textbf{Specialist Methods:}}} \\
      MattNet~[CVPR'18]~\cite{mattnet} & ResNet-101& Bi-LSTM & 48.24 & 47.51 & 41.15 & 59.30 & 58.66 & 44.04 & 46.14 & 45.33 & 41.32 \\
      CRIS~[CVPR'22]~\cite{cris} & CLIP-ResNet-101 & CLIP-B & 56.27 & 55.34 & - & 63.42 & 63.82 & - & 51.79 & 51.04  & - \\
      LAVT~[CVPR'22]~\cite{lavt} & Swin-B & BERT-B & 58.40 & 57.64 & 49.32 & 65.90 & 65.32 & 49.25 & 55.83 & 55.04 & 48.46 \\
      ReLA~[CVPR'23]~\cite{rela} & Swin-B & BERT-B & 63.60 & 62.42 & 56.37 & 70.03 & 69.26 & 59.02 & 61.02 & 59.88 & 58.40 \\
      EEVG~[ECCV'24]~\cite{eevg} & ViT-B & BERT-B & 62.75 & 64.04 & - & 70.93 & 71.65 & - & 62.79 & 62.77 & - \\
      HieA2G~[AAAI'25]~\cite{hieA2G} & Swin-B & RoBERTa-B & 68.40 & 64.20 & 62.80  & 72.00 & 70.40 &  63.40  & 62.80 & 60.80 & 61.00\\ 
      \hline
      \rowcolor{gray!10} 
      \textbf{DeRIS-B (Ours)} & Swin-S & BEiT3-B & \underline{74.10} & \underline{68.06} & \underline{77.03} & {73.72} & {71.99} & \underline{75.98} & \underline{65.63} & \underline{64.65} & \underline{63.44} \\
      \rowcolor{gray!10}
      \textbf{DeRIS-L (Ours)} & Swin-B & BEiT3-L & \textbf{77.67} & \textbf{72.00} & \textbf{82.22} & \textbf{75.30} & \textbf{73.73} & \textbf{78.30} & \textbf{67.99} & \textbf{67.38} & \textbf{66.81} \\
      \specialrule{.1em}{.05em}{.05em}
      \end{tabular}
  }
  \label{tab:sota_gres}
  \vspace{-5pt}
\end{table*}

\section{Experiments}
\subsection{Settings}
\label{sec:setting}
We conduct extensive experiments on four publicly available datasets: RefCOCO/+/g~\cite{refcoco,refcocog-umd}, gRefCOCO~\cite{rela}. For the perception branch, we initialize the model using pretrained weights from Mask2Former~\cite{mask2former} and use an input resolution of 384$\times$384. For the cognition branch, we initialize the model using pretrained weights from BEiT3~\cite{beit3} and use an input resolution of 224$\times$224. Additional implementation details are provided in the Appendix C.

\subsection{Main Results}
\noindent\textbf{(1) RIS Results:}  
Table~\ref{tab:sota_ris} presents a comparative evaluation of DeRIS against recent SOTA methods on the RefCOCO/+/g~\cite{refcoco,refcocog-umd} datasets. Notably, DeRIS surpasses existing advanced models by a substantial margin. Specifically, compared to the recent specialist method OneRef-L~\cite{oneref}, DeRIS-L achieves improvements of 4.46\%, 4.68\%, and 4.33\% in mIoU on the validation sets of RefCOCO, RefCOCO+, and RefCOCOg, respectively. Furthermore, we extend DeRIS to a large-model variant by integrating Qwen2-OV-7B~\cite{llava-onevision}, replacing BEiT3~\cite{beit3} with a more powerful multi-modal large language model while employing LoRA~\cite{lora} for efficient training. Experimental results demonstrate that, at a comparable model scale, DeRIS-7B outperforms SAM4MLLM-8B~\cite{sam4mllm} by 4.25\%, 5.70\%, and 5.12\% in cIoU on the validation sets of RefCOCO/+/g datasets.

\noindent\textbf{(2) GRES Results:}  
Table~\ref{tab:sota_gres} summarizes the performance of DeRIS on the gRefCOCO~\cite{rela} dataset, where it consistently outperforms prior methods across nearly all metrics, with a particularly significant improvement in the N-acc metric. These results highlight DeRIS’s robustness in handling both non-referent and multi-referents scenarios, demonstrating its strong generalization capabilities.

\begin{table}
  \centering
  \footnotesize
  \setlength{\tabcolsep}{6.0pt}
  \renewcommand\arraystretch{1.2}
  \caption{The analysis of the perception and cognition capabilities. The results are conducted on the gRefCOCO~\cite{rela} dataset.}
  \vspace{-5pt}
  \resizebox{0.95\linewidth}{!}
  {
  \begin{tabular}{cc|ccc}
  \specialrule{.1em}{.05em}{.05em} 
  Cognition & Perception & N-acc. & gIoU & cIoU \\
  \hline
  BERT-B      &   \multirow{4}{*}{Swin-S} & 66.13 & 64.55 & 56.26  \\
  ViLT-B      &      & \cellcolor{gray!5 }68.25~\increase{2.12} & \cellcolor{gray!5 }67.12~\increase{2.57} & \cellcolor{gray!5 }61.23~\increase{4.97} \\
  BEiT3-B      &      & \cellcolor{gray!8 }70.25~\increase{4.12} & \cellcolor{gray!8 }70.64~\increase{6.09} & \cellcolor{gray!8 }66.31~\increase{10.05} \\
  BEiT3-L      &      & \cellcolor{gray!10 }73.16~\increase{7.03} & \cellcolor{gray!10 }73.53~\increase{8.98} & \cellcolor{gray!10 }69.14~\increase{12.88}  \\
  \hline
  \hline
  \multirow{3}{*}{BEiT3-B}    &  Swin-T    & 70.22 & 70.13 & 65.54 \\
                               &  Swin-S    & \cellcolor{gray!5 }70.25~\increase{0.03} & \cellcolor{gray!5 }70.64~\increase{0.51} & \cellcolor{gray!5 }66.31~\increase{0.83} \\
                               &  Swin-B    & \cellcolor{gray!10 }70.35~\increase{0.13} & \cellcolor{gray!10 }70.85~\increase{0.72} & \cellcolor{gray!10 }66.78~\increase{1.20} \\
  \specialrule{.1em}{.05em}{.05em}
  \end{tabular}
  }
  \label{tab:ab_recognition_perception}
  \vspace{-5pt}
\end{table}

\begin{table}
  \centering
  \footnotesize
  \setlength{\tabcolsep}{4.0pt}
  \renewcommand\arraystretch{1.2}
  \caption{The analysis of cognition component capabilities, including comparisons with MLLM. The metrics are based on the average gIoU across several subsets.}
  \vspace{-5pt}
  \resizebox{1.0\linewidth}{!}{
  \begin{tabular}{l|c|ccc}
  \specialrule{.1em}{.05em}{.05em} 
  Cognition & Perception & RefCOCO & RefCOCO+ & RefCOCOg-U \\
  \hline
  BEiT3-0.2B~\cite{beit3} & \multirow{4}{*}{Swin-B} &  79.06 &72.43 & 73.98  \\
  Qwen2-0.5B~\cite{llava-onevision} & & \cellcolor{gray!5 }81.32~\increase{2.26} & \cellcolor{gray!5 }74.63~\increase{2.20} & \cellcolor{gray!5 }74.96~\increase{0.98} \\
  Mipha-3B~\cite{mipha} & & \cellcolor{gray!8 }82.16~\increase{3.10} & \cellcolor{gray!8 }79.19~\increase{6.76} & \cellcolor{gray!8 }77.38~\increase{3.40} \\
  Qwen2-7B~\cite{llava-onevision} & & \cellcolor{gray!10 }82.87~\increase{3.81} & \cellcolor{gray!10 }79.88~\increase{7.45} & \cellcolor{gray!10 }78.30~\increase{4.32} \\
  \specialrule{.1em}{.05em}{.05em}
  \end{tabular}
  }
  \label{tab:ab_cognition}
  \vspace{-10pt}
\end{table}

\subsection{Quantitative Analyses}
\label{sec:quantitative_analysis}
In this section, we present a series of analyses from a quantitative perspective. Comprehensive qualitative analyses are provided in Appendix F.
\subsubsection{Influence of Cognition and Perception Capability}
\label{sec:ab_cognition_perception}

\noindent\textbf{(1) Cognition Capability is Crucial for RIS.}
The decoupled structure, which separates the RIS task into cognition and perception branches, provides a robust foundation for analyzing the relative impact of each component on model performance. Specifically, the cognition branch is tasked with determining whether the perceptual output aligns with the referent described in the input sentence by interpreting vision-text semantics. We conduct a fair comparison by systematically varying the model scales of the cognition and perception branches, as presented in Table~\ref{tab:ab_recognition_perception}.
In the first ablation group of Table~\ref{tab:ab_recognition_perception}, we observe a nearly 10-point improvement in cIoU when transitioning from the BERT-B~\cite{bert} model to the BEiT3-B~\cite{beit3} model. Further scaling the BEiT3 model from base to large yields an additional significant 3-point gain in cIoU. In contrast, the perception branch exhibits minimal performance improvement with increasing model size. These findings lead us to conclude that, for current RIS tasks, cognitive ability plays a more critical role than perceptual ability.

\noindent\textbf{(2) Cognitive Ability as the Primary Bottleneck with Significant Room for Improvement.}  
As shown in Table~\ref{tab:ab_cognition}, the model's performance steadily improves as the scale of the cognitive model increases. A similar trend is observed in MLLM~\cite{Qwen2VL,mipha}, further confirming that the cognitive model's capability serves as the primary bottleneck in RIS tasks. This suggests that enhancing multi-modal understanding is the key to further improving RIS performance.  

\subsubsection{Effectiveness of Fine-Grained Segmentation}

\noindent\textbf{(1) Fine-Grained Segmentation Analysis.} 
DeRIS decouples perception and cognition, effectively leveraging the strengths of both, which leads to improvements in both perceptual and cognitive capabilities. We evaluate the model's fine-grained segmentation ability using the Pr@0.9, with results shown in Table~\ref{tab:ab_finegrained_perception}. It demonstrates that DeRIS significantly enhances fine-grained segmentation performance.

\noindent\textbf{(2) Efficiency Analysis.} 
Cognition-centric methods typically employ the vision transformer architecture, which has a quadratic computational complexity with respect to the image resolution, making them inefficient for processing high-resolution inputs. In contrast, perception-centric methods use hierarchical architectures, thus enabling more efficient processing of high-resolution inputs. As shown in Table~\ref{tab:ab_finegrained_perception}, our method maintains high inference efficiency even with powerful performance.

\begin{table}
  \centering
  \footnotesize
  \setlength{\tabcolsep}{4.0pt}
  \renewcommand\arraystretch{1.2}
  \caption{The effectiveness and efficiency of different architecture focuses under varying resolutions. Infer Time is measured on a single NVIDIA A100 GPU.}
  \vspace{-5pt}
  \resizebox{0.9\linewidth}{!}
  {
    \begin{tabular}{l|ccc}
    \specialrule{.1em}{.05em}{.05em} 
    Structure & gIoU  & Pr@0.9 & Infer Time (ms) \\
    \hline
    \multicolumn{4}{c}{\textbf{Resolution=224}}\\
    Cognition-centric   & 68.54  & 8.32 & 14.7 \\
    Perception-centric  & 64.55~\decrease{3.99}  & 15.79~\increase{7.47} & 11.2~\increase{24\%}  \\
    \rowcolor{gray!10} Integrated  & 70.64\increase{2.10}  & 20.93\increase{12.61} & 15.9\decrease{8\%}\\
    \hline
    \multicolumn{4}{c}{\textbf{Resolution=384}}\\
    Cognition-centric  & 69.87 & 13.98 & 28.1 \\
    Perception-centric  & 67.12~\decrease{2.75}  & 23.08~\increase{9.10} & 20.0~\increase{29\%} \\
    \rowcolor{gray!10}  Integrated  & 71.74~\increase{1.87}  & 28.39~\increase{14.41} & 22.7~\increase{19\%} \\
    \specialrule{.1em}{.05em}{.05em}
    \end{tabular}
  }
  \label{tab:ab_finegrained_perception}
  \vspace{-5pt}
\end{table}

\subsubsection{Analysis of Non-referent Judgement}
\label{sec:ab_non_referent}
\noindent\textbf{(1) Influence of Non-referent Prediction During Training and Testing.}  
The instance-level prediction framework does not explicitly require non-referent classification, as the model inherently assumes the absence of a target when all object queries yield low confidence scores. However, we observe that incorporating auxiliary supervision for non-referent samples during training significantly improves the model’s ability to identify such cases.  
Moreover, in the post-processing stage, multiplying $\mathcal{P}_{nr}$ with $\mathcal{P}_{ref}$ during inference results in a notable performance gain, as shown in Table~\ref{tab:ab_target_existence}. This indicates that both implicit target existence learning during training and explicit target presence verification during inference contribute to performance improvements. These findings suggest that the model exhibits an inherent instability in global target existence prediction.

\noindent\textbf{(2) Impact of Non-referent Sample Conversion.}  
In the gRefCOCO~\cite{rela} dataset, non-referent samples constitute approximately 9\% of the total data, making non-referent classification a long-tail distribution problem. To mitigate this issue, we introduce NSC augmentation , a strategy that dynamically expands the diversity of non-referent samples during training. As shown in Table~\ref{tab:ab_target_existence}, this approach results in more than 15-point improvement in N-acc. Additionally, NSC enables the model to achieve comparable performance without the need to introduce $\mathcal{P}_{nr}$ during post-processing, further streamlining inference.

\begin{figure}
  \centering
  \includegraphics[width=0.9\linewidth]{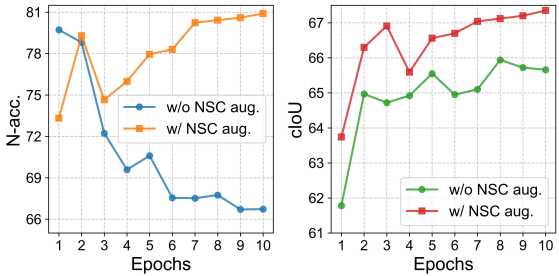}
  \vspace{-10pt}
  \caption{Impact of NSC on training stability and convergence.}
  \vspace{-10pt}
  \label{fig:nsc_stability}
\end{figure}

\noindent\textbf{(3) Stability Analysis of the Training Process.}
As shown in Fig.~\ref{fig:nsc_stability}, we observe that the N-acc metric fluctuates significantly during training on the gRefCOCO~\cite{rela} dataset, which leads to instability in other metrics such as cIoU. We attribute this instability primarily to the limited diversity of non-referent image-text samples in the dataset, which causes the model to overfit. However, after introducing the NSC strategy, the N-acc metric becomes more stable throughout the training process. This indicates that the NSC strategy effectively alleviates the issue of non-referent sample scarcity, improving the model's generalization ability.

\begin{table}
  \centering
  \footnotesize
  \setlength{\tabcolsep}{6.0pt}
  \renewcommand\arraystretch{1.2}
  \caption{The impact of non-referent supervision and judgement, as well as the influence of incorporating the NSC strategy.}
  \vspace{-5pt}
  \resizebox{0.85\linewidth}{!}{
  \begin{tabular}{cc|ccc}
  \specialrule{.1em}{.05em}{.05em} 
  Train & Test&  N-acc & gIoU & cIoU  \\
  \hline
  \multicolumn{5}{c}{\textbf{w/o NSC Augmentation}}\\
  & & 60.19 & 66.09 & 63.98  \\
  \rowcolor{gray!5} $\checkmark$ &  & 62.71\increase{2.52} & 67.73\increase{1.64} & 65.31\increase{1.33}  \\
  \rowcolor{gray!10} $\checkmark$ & $\checkmark$ & 68.80\increase{6.09} & 70.10\increase{2.37} & 66.40\increase{1.09}  \\
  \hline
  \multicolumn{5}{c}{\textbf{w/ NSC Augmentation}}\\
  & & 75.36\textbf{\increase{15.17}} & 71.82\textbf{\increase{5.73}} & 66.33\textbf{\increase{2.35}} \\
  \rowcolor{gray!5} $\checkmark$ & & 79.25\increase{3.89} & 73.77\increase{1.95} & 67.49\increase{1.16} \\
  \rowcolor{gray!10} $\checkmark$ & $\checkmark$ & 79.32\increase{0.07} & 72.78\decrease{0.99} & 66.41\decrease{1.08} \\
  \specialrule{.1em}{.05em}{.05em}
  \end{tabular}
  }
  \label{tab:ab_target_existence}
  \vspace{-5pt}
\end{table}

\subsection{Ablation Study}
\label{sec:ablation_study}

\subsubsection{Synergy Forms between Perception and Cognition}
\label{sec:ab_loopback}
The integration of perception and cognition branches is a pivotal aspect of our decoupled framework. To investigate this, we evaluated four interaction forms: P-to-C (Baseline), C-to-P, Hierarchical Combined, and Loopback Synergy, with results presented in Table~\ref{tab:ab_loopback}. Our findings indicate that transferring object queries from the cognition branch to the perception branch notably decelerates model convergence. We attribute this to the perception branch's optimization being constrained by the cognition pipeline. Conversely, the reverse flow (P-to-C) supports typical convergence rates. The Hierarchical Combined approach, which fuses $f_v$ with $f_{1-3}$ at the feature level using SimFPN~\cite{vitdet}, offers modest performance improvements but substantially reduces training efficiency. The proposed loopback synergy strategy, leveraging multi-rounds iterative object query exchanges between branches to enhance both cognitive and perceptual capabilities. This method improves performance by 1-2 points without adding notable computational cost.

\begin{table}
  \centering
  \footnotesize
  \setlength{\tabcolsep}{6.0pt}
  \renewcommand\arraystretch{1.2}
  \caption{The impact of different connection structures.}
  \vspace{-5pt}
  \resizebox{1.0\linewidth}{!}
  {
    \begin{tabular}{l|ccc}
    \specialrule{.1em}{.05em}{.05em} 
    Structure & gIoU  & cIoU & Train Time (h) \\
    \hline
    Query Transfer: P-to-C  &  69.98 & 65.49  &  2.7 \\
    Query Transfer: C-to-P & 56.77\decrease{13.21}  & 54.80\decrease{10.69} &  2.7\decrease{0\%} \\
    Hierarchical Combined & 70.13\increase{0.15}  & 66.32\increase{0.83} & 3.2\decrease{18\%}\\
    \rowcolor{gray!10}  Loopback Synergy& 71.37\increase{1.39}  & 67.27\increase{1.78} & 2.8\decrease{3\%}\\
    \specialrule{.1em}{.05em}{.05em}
    \end{tabular}
  }
  \label{tab:ab_loopback}
  \vspace{-5pt}
\end{table}

\begin{table}
  \centering
  \footnotesize
  \setlength{\tabcolsep}{6.0pt}
  \renewcommand\arraystretch{1.2}
  \caption{The impact of non-referent sample conversion rate $R_c$.}
  \vspace{-5pt}
  \resizebox{0.73\linewidth}{!}{
  \begin{tabular}{c|ccc}
  \specialrule{.1em}{.05em}{.05em} 
  $ R_c $ & N-acc. & gIoU & cIoU \\
  \hline
  0\%   &70.25 & 70.64 & 66.31 \\
  5\%   & 74.48\increase{4.23}   & 72.49\increase{1.85} & 67.22 \increase{1.91} \\
  10\%  & 75.46\increase{0.98}  & 72.50\increase{0.01} & 66.76\decrease{0.46}  \\
  \rowcolor{gray!10} 15\%  & 79.25 \increase{3.79} & 73.77\increase{1.27} & 67.49\increase{0.73}  \\
  20\%  & 79.39\increase{0.14}   & 73.67\decrease{0.10} & 67.07\decrease{0.42} \\
  \specialrule{.1em}{.05em}{.05em}
  \end{tabular}
  }
  \label{tab:ab_R_c}
  \vspace{-5pt}
\end{table}

\subsubsection{Impact of Conversion Rate of Non-referent}
\label{sec:ab_R_c}

The proposed NSC data augmentation strategy transforms image-sentence pairs from referent to non-referent cases. The conversion probability $R_c$ serves as a critical hyperparameter in this process. As shown in Table~\ref{tab:ab_R_c}, we observe that increasing $R_c$ leads to a steady improvement in non-referent accuracy (N-acc). However, when $R_c$ exceeds a certain threshold, performance on other metrics, such as cIoU, begins to decline. This occurs primarily because enhancing the diversity of non-referent samples prompts the model to assign higher confidence to predicting the absence of a target, effectively mitigating the low-confidence issue in tail categories stemming from the long-tail distribution of the dataset. In this paper, we set $R_c$=15\% to address the limited diversity and scarcity of non-referent samples.

\section{Conclusion}

In this paper, we introduce \textbf{DeRIS}, a novel framework for RIS that decouples perception and cognition to harness their complementary strengths. Our analyses reveal that while existing models are primarily limited by the inadequate contextual image-text understanding. To address this gap, we propose a multi-round loopback synergy mechanism that iteratively enhances cognitive and perceptual capabilities through query refinement. Additionally, we present a non-referent sample conversion strategy to boost confidence in non-referent judgment, mitigating the effects of long-tail distribution in the dataset that contribute to uncertainty in handling non-referent cases. Extensive experiments demonstrate that our approach significantly enhances performance across both RIS and GRES tasks.

\section*{Acknowledgements}
This work is supported by the National Natural Science
Foundation of China under Nos. 62276061 and 62436002.
This work is also supported by Research Fund for Advanced Ocean Institute of Southeast University (Major Program MP202404). This work is also supported by the SEU Innovation Capability Enhancement Plan for Doctoral Students (CXJH\_SEU25125).

{\small
\bibliographystyle{ieeenat_fullname}
\bibliography{DeRIS}
}

\end{document}